\documentclass[acmsmall,screen,nonacm]{acmart}

\usepackage{amsmath}
\usepackage{subfig}
\usepackage{xspace}
\usepackage{stmaryrd}
\usepackage{tikz}
\usetikzlibrary{arrows,shapes,positioning,matrix}
\usetikzlibrary{intersections}
\usetikzlibrary{calc}
\usetikzlibrary{patterns}
\usetikzlibrary{trees,snakes}
\usetikzlibrary{shadows,decorations.pathmorphing,decorations.markings}
\usetikzlibrary{fit}
\tikzstyle{every picture}+=[remember picture]
\tikzstyle{na} = [baseline=-.5ex]

\usepackage{hyperref}

\usepackage{todonotes}
\usepackage{pgfplots}
\usepackage{listings,lstlinebgrd}
\usepackage{xcolor}
\usepackage{comment}
\definecolor{purpy}{rgb}{0.769,0.02,0.894} 
\definecolor{bluepy}{rgb}{0.082,0.02,1}
\definecolor{brownpy}{rgb}{0.557,0.388,0.184}
\definecolor{greenpy}{rgb}{0.118,0.553,0.388}
\usepackage{caption}
\DeclareCaptionFormat{listing}{\rule{\dimexpr\linewidth\relax}{0.4pt}\par\vskip1pt#1#2#3}
\graphicspath{fig/}
\usepackage{appendix}
\usepackage{float}
\usepackage{algorithm}
\usepackage[noend]{algorithmic}

\newcommand{\revision}[1]{{#1\xspace}}

\AtBeginDocument{%
  }
\newcommand{\ema}[1]{\ensuremath{#1}\xspace}
\newcommand{\abprio}{\ema{\texttt{ab-Baselines}}}
\newcommand{\abwf}{\ema{\texttt{ab-WorstFit}}}
\newcommand{\abff}{\ema{\texttt{ab-FirstFit}}}
\newcommand{\abbf}{\ema{\texttt{ab-BestFit}}}

\newcommand{\unifDef}{\ema{\textsc{Uniform}(20,100)}}
\newcommand{\weibDef}{\ema{\textsc{Weibull}(3.0,45)}}

  \newcommand{\bestfit}{{\texttt BestFit}\xspace}
  \newcommand{\worstfit}{{\texttt WorstFit}\xspace}
  \newcommand{\firstfit}{{\texttt FirstFit}\xspace}

  \newcommand{\cTwelve}{{\texttt c12}\xspace}
  \newcommand{\smoothcTwelve}{{\texttt Smooth\_c12}\xspace}
  \newcommand{\cFourteen}{{\texttt c14}\xspace}
  \newcommand{\eoh}{{\texttt EoH}\xspace}

\newcommand{\romera}{Romera et al.~[\citeyear{romera2024mathematical}]\xspace}
\newcommand{\romeras}{Romera et al.'s\xspace}

\setcopyright{cc}
\setcctype{by}
\copyrightyear{2026}
\acmYear{2026}
\acmDOI{XXXXXXX.XXXXXXX}

\title{An In-depth Study of LLM Contributions to the Bin Packing Problem}
\author{Julien Herrmann}
\affiliation{%
  \institution{CNRS-IRIT}
  \city{Toulouse}
  \country{France}
}
\email{julien.herrmann@irit.fr}
\orcid{0000-0003-4935-2368}

\author{Guillaume Pallez}
\affiliation{%
  \institution{Inria}
  \city{Rennes}
  \country{France}
}
\email{guillaume.pallez@inria.fr}
\orcid{0000-0001-8862-3277}

\ccsdesc[500]{Computing methodologies~Heuristic function construction}
\ccsdesc[300]{Theory of computation~Design and analysis of algorithms}
\begin{abstract}

Recent studies have suggested that Large Language Models (LLMs) could provide interesting ideas contributing to
\emph{mathematical discovery}. This claim was motivated by reports that 
LLM-based genetic algorithms produced heuristics offering new insights into the online
bin packing problem under uniform and Weibull
distributions.
In this work, we reassess this claim through a detailed analysis of the heuristics produced by LLMs,
examining both their behavior and interpretability.
Despite being human-readable, these heuristics remain largely opaque even to domain experts.
Building on this analysis, we propose a new class of algorithms tailored to these specific bin packing instances.
The derived algorithms are significantly simpler, more efficient, more interpretable, and more generalizable,
suggesting that the considered instances are themselves relatively simple.
We then discuss the limitations of the claim regarding LLMs' contribution to
this problem, which appears to rest on the mistaken assumption that the instances 
had previously been studied.
Our findings instead emphasize the need for rigorous validation and contextualization 
when assessing the scientific value of LLM-generated outputs.
\end{abstract}

\begin{document}

\maketitle


\section{Introduction}

Heuristics and approximation algorithms are widely used to tackle NP-hard combinatorial optimization problems arising in real-world applications.
Manually designing, adapting, and tuning an effective heuristic for a given problem demand expert experience and a deep understanding of the problem's constraints and objectives.
Hyper-heuristics have therefore emerged as an active research area, aiming to automate algorithm design by selecting and tuning effective heuristics from a set of heuristic components~\cite{pillay2018hyper, stutzle2018automated}.
Genetic algorithms, for instance, evolve a pool of candidate solutions through
crossover and mutation \revision{operators~\cite{koza1990genetic}}.
However, the intrinsic complexity and heterogeneity of NP-hard problems
usually limit these operators to narrow classes \revision{of code space of functions}, typically
parameterized operations predefined by human experts~\cite{pillay2018hyper}.

Recent work~\cite{romera2024mathematical,bengio2021machine, wu2024evolutionary}
has proposed using Large Language Models (LLMs) to open up new possibilities for
the genetic operators, enabling (i) exploration of different heuristic
spaces; and (ii) a better understanding of the resulting solution. These approaches
combine LLMs with evolutionary search techniques to produce heuristics in an
automated manner.
One of the most visible recent papers, \textit{``Mathematical discoveries from
program search with large language models''}~\cite{romera2024mathematical},
published in \textit{Nature} in December 2023 and cited by \revision{744 papers
(according to Google Scholar) as of October 1st 2025}~\footnote{\revision{At the time of
the final version of this article (June 10, 2026), this number had increased to 1498
citations.}}, introduces the FunSearch
procedure. It models heuristic design as a search problem in the space of
functions, and uses LLMs in an evolutionary framework to iteratively improve the
quality of the generated functions. FunSearch has been applied to three problems:
\textit{cap set}~\cite{grochow2019new}, \textit{admissible set}~\cite{edel2004extensions}, and \textit{online bin
packing}~\cite{coffman1984approximation}.

The key claims of \romera are that
\textit{``FunSearch outputs programs that tend to be interpretable''} and that
\textit{``LLMs should be seen as a source of diverse (syntactically correct)
programs with occasionally interesting ideas''}.
\revision{However, no work has been conducted to analyse the generated heuristics.
Their performance are only accessed by comparing with some baseline heuristics on the data distribution they have been evolved.}
\revision{In this paper, we use ``interesting ideas'' in a deliberately stronger
sense than mere empirical improvements on a benchmark. We consider an idea
scientifically interesting when it provides insight into why a method works, can
be communicated independently of the generated code, and can potentially guide
the design of new methods beyond the original training setting. By contrast, we
use ``mathematical discovery'' to refer to a stronger claim, requiring not only
improved performance but also evidence of novelty, significance, and conceptual
or theoretical understanding. In this manuscript, we reassess the key claims of
\romera, focusing on the {\em online bin packing}
problem. {\bf Can the contributions of the LLM-based evolutionary framework for this specific NP-hard problem be
considered scientifically interesting in this sense, and how should they be evaluated?}
While the generated heuristics are straightforward to implement and benchmark, their claimed interpretability remains questionable, as subsequent work
continues to rely on LLM-based generation rather than identifying and exploiting the underlying principles that could be extracted from \romera.
We first show that interpreting the proposed heuristics is indeed nontrivial.} We
then show that this difficulty is avoidable by providing a simple,
more efficient, two-parameter heuristic that generalizes to a much broader
set of instances.

The rest of the paper is organized as follows. We first review related work 
on hyper-heuristics and LLM-assisted algorithm design in
Section~\ref{sec:related}. We then discuss the interpretability of the heuristics proposed in \romera 
and subsequent work in Section~\ref{sec:interpretability}, before discussing
their scientific contribution in Section~\ref{sec:interesting}. In particular we
propose and evaluate a new, two-parameter simple heuristic for these instances.
Building on these observations, we critically reassess the claim that LLM-based evolutionary frameworks can
occasionally contribute scientifically interesting ideas, and argue that, for bin packing, 
the available evidence in \romera is insufficient to establish this claim.


\section{Related work}
\label{sec:related}

There is a long history of {\em Automatic Heuristic Algorithm} design, also known as
{\em hyper-heuristics}, for combinatorial
problems~\cite{burke2003hyper,burke2013hyper,stutzle2018automated}. This body of work
includes hyper-heuristics for the bin packing
problem~\cite{ross2002hyper,sim2015lifelong}.
Among these approaches, genetic programming~\cite{back1997handbook, eiben2015evolutionary}
is a generic optimization principle inspired by natural evolution that provides
an interpretable approach to algorithm design~\cite{mei2022explainable,
jia2022learning}. However, it relies on handcrafted algorithmic components and
domain knowledge, and is typically constrained by the code space of functions predefined by
human experts.

\paragraph{On the inclusion of LLMs in the design of hyper-heuristics}

To this date, standalone LLMs, even when aided by prompt engineering, remain
limited by hallucinations and unreliable reasoning, making them insufficient for solving
open problems~\cite{bang2023multitask}. To
overcome these limitations, recent work has combined them with evolutionary
frameworks~\cite{wu2024evolutionary, hemberg2024evolving, wang2025large}.
It is believed that LLMs could leverage the definition of genetic operators for
code generation within evolutionary frameworks~\cite{liventsev2023fully,
ma2023eureka, lehman2023evolution}. The crossover and
mutation operators when evolving heuristics are handled by LLMs that have been
fine-tuned on a large corpus of code, broadening the heuristic search space.

\paragraph{Romera-Paredes et al.' paper}

One of the most influential works exploring this idea is \romera.
The authors introduce FunSearch, a distributed evolutionary framework that
combines a frozen LLM as a code generator with an evaluation function that scores
candidate heuristics.
It evolves the critical parts of a predefined heuristic skeleton using best-shot
prompting, island-based population management, and asynchronous
sampling/evaluation to iteratively improve the quality of the generated heuristics.
FunSearch is based on Codey, an LLM from the PaLM 2 model
family~\cite{anil2023palm}, and requires millions of samples (i.e., queries to the
LLM) and days of training to design a heuristic suited to a given dataset. 
\romera focus on three
NP-hard combinatorial problems, including the \textit{online bin
packing}~\cite{coffman1984approximation} problem. They claim that FunSearch
allows ``the discovery of new scientific results'' but no in-depth analysis 
of the generated heuristics is provided to establish whether the discoveries are
truly new.

\paragraph{Implications of Romera-Paredes et al.' work}

At the time of the writing (October 1st 2025), \cite{romera2024mathematical} has
been cited by 744 research papers. Only
91 of them specifically mention \textit{bin packing}. An evaluation of a convenience
sample biased toward the most cited of the remaining 653 papers suggests
that most of them cite the article primarily by title to assert that coupling LLMs with
evolutionary frameworks can yield ``mathematical discoveries''.
Among the 91 papers that cite Romera-Paredes and mention \textit{bin packing}, 
\revision{some work on the bin packing problem}. Among those, two main categories emerge:
(i) papers that design improved LLM-based
hyper-heuristics~\cite{liu2024evolution,dat2025hsevo,ye2024reevo}—for instance
refining code generation via more suitable evolutionary frameworks and mitigating
FunSearch's inefficiency due to millions of LLM calls; and
(ii) papers that evaluate the claimed efficiency~\cite{sim2025beyondhype}.
{\em Notably, although \romera
claim that their examples demonstrate that program search with LLMs can enable new
mathematical discoveries, at this time, we have not found work that builds on
their generated heuristics.}

The {\em Evolution of Heuristics} (EoH) framework~\cite{liu2024evolution}
jointly evolves high-level natural language
descriptions (\textit{thoughts}) and corresponding code implementations using LLMs
within a population-based evolutionary algorithm. The framework applies prompt-based
variation operators to generate and refine heuristics for NP-hard combinatorial
problems, including bin packing.
ReEvo~\cite{ye2024reevo} and HSEvo~\cite{dat2025hsevo} extend
FunSearch by incorporating reflective feedback loops, hybrid evolutionary
operators, and diversity-preserving mechanisms. 
MEoH~\cite{yao2025multi} extends the single-objective approach of FunSearch 
to the multi-objective setting, considering additional practical criteria beyond 
heuristic performance, and uses FunSearch as a baseline.

Sim et al.~[\citeyear{sim2025beyondhype}] conduct an extensive benchmarking study on
LLM-evolved heuristics for bin packing. Their experiments reveal that many
LLM-evolved heuristics tend to overfit and generalize poorly
beyond narrow instance clusters. This critique underscores the need for
standardized benchmarking and caution in claiming broad applicability. 

However, to the best of our knowledge, 
no studies have examined the interpretability of LLM-evolved
heuristics for the bin packing problem to understand \textit{why} they
outperform baseline heuristics and whether they constitute ``truly new discoveries''.
Likewise, no study has instantiated the insight attributed to the
LLM–evolutionary pipeline as a reusable algorithmic principle for bin packing.

\paragraph{Stochastic Online bin packing / Average analysis on online bin packing}

When claiming a mathematical discovery, it is essential to review the relevant state of
the art— something that was not done in \romeras work. The problem of online bin
packing, where item sizes are independent and identically distributed (i.i.d) random variables drawn from
a stochastic distribution, is known as {Stochastic Online Bin
Packing}~\cite{johnson1974worst,ayyadevara2025near}, or, equivalently,
as the \emph{average case-analysis} of online bin packing~\cite{shor1986average}. For such
algorithms, performance is generally assessed through the \emph{expected competitive
ratio} and the \emph{expected wasted space}, where wasted space denotes the total amount
of unused capacity in partially filled bins.
Most existing studies assume that the underlying distribution is unknown and, therefore, cannot be 
leveraged by the algorithm~\cite{ayyadevara2025near,shor1986average}. 

In contrast, \romera explicitly consider a known distribution that the algorithm can exploit. 
When such information is available, prior research has focused almost exclusively on the uniform case
—either $\mathcal{U}[0,1]$~\cite{coffman1980stochastic,coffman1997bin,shor1986average} 
or $\mathcal{U}[0,b]$ with $b \leq 1$~\cite{bentley1984some}. In the latter setting,
Bentley et al.~\cite{bentley1984some} showed that the \emph{first-fit} and \emph{best-fit} algorithms are
asymptotically optimal and achieve constant wasted space
with high probability when $b \leq 0.5$.
To the best of our knowledge, there is no prior research on the stochastic online bin packing
problem where the uniform distribution admits a positive lower bound $a>0$, nor any that considers
non-uniform distributions such as the Weibull distribution,
as explored in \romera.


\section{Interpretability of LLM-evolved heuristics}
\label{sec:interpretability}

One of the key claims of \romera is that
\textit{``FunSearch outputs programs that tend to be interpretable''}. 
In this section we focus on the technical content of \romera through an in-depth analysis of the produced
heuristics. Our goal is to assess the interpretability of the reported discovery. We further extend
this analysis by considering subsequent work, notably Liu et al.~\cite{liu2024evolution}.

All the code used for the experiments and analyses in this paper is
available at {\em  \href{https://gitlab.inria.fr/ai-interpretability/llm-4-bin-packing}{https://gitlab.inria.fr/ai-interpretability/llm-4-bin-packing}.}

\subsection{Online bin packing and priority functions}

\revision{{\em Bin packing}~\cite{coffman1984approximation} is the problem of
packing a set of items of various sizes into bins. Each bin
has the same fixed capacity that cannot be exceeded by the sum of the sizes of
the items assigned to it. The objective is to minimize the total number of bins
used.}
It is a well-known NP-hard combinatorial problem with applications in many
areas, from operations research to scheduling jobs on compute clusters. In the
{\em online} scenario, items are packed as they arrive, in contrast with
the {\em offline} scenario where all items are known beforehand.
Solving {\em online bin packing} problems therefore requires heuristics that greedily decide 
to which bin each incoming item should be assigned.

A common approach to solving the online bin packing problem is to use
priority functions: at all times, bins are sorted according to a {\em priority function} that
depends on their remaining space and the size of the incoming item. The heuristic then
allocates the item to the bin with the highest priority score.

The most widely used heuristics for {\em bin packing} are \firstfit, \bestfit,
and \worstfit. 
In \firstfit, bins are prioritized by decreasing index.
In \bestfit (resp. \worstfit), bins are prioritized by decreasing
(resp. increasing) remaining available space.
These simple heuristics show good performance in practice~\cite{kenyon1996best}
both in the worst case (or adversarial case) and in the average uniform case~\cite{bentley1984some}.

\romera and subsequent works considered
special instances of {\em online bin packing} instances where incoming item sizes
follow two probability distributions: \unifDef and \weibDef. For each of these
instances they developed specific heuristics which we analyze in
Section~\ref{sec:unifdef} and Section~\ref{sec:weibdef}. Subsequent
work~\cite{sim2025beyondhype} has shown that these heuristics fail to generalize across
different parametrization of the problem.

\subsection{Item sizes following a \unifDef distribution with bin capacity 150}
\label{sec:unifdef}
We begin by discussing the heuristic developed for the \unifDef problem.
The instances consist of items with integer sizes drawn from a
uniform distribution in $\llbracket 20,100 \rrbracket$. Each bin has a capacity of 150.

\paragraph{Heuristic generation}
The best heuristic generated by \romera for \unifDef is denoted \cTwelve
(Algorithm~\ref{alg:c12}). It was generated by evolving FunSearch on 20 training
instances, randomly generated using the same parameters as those in the first
of the OR-Library bin packing benchmarks, OR1~\cite{beasley1990or}. The training
instances consist of 120 items.

\begin{center}
\begin{lstlisting}[language=Python,caption=\cTwelve~\cite{romera2024mathematical} for items of distribution \unifDef,label=alg:c12]
def priority_c12(item: float, bins: np.ndarray) -> np.ndarray:
    def s(bin, item):
        if bin - item <= 2:
            return 4
        elif (bin - item) <= 3:
            return 3
        elif (bin - item) <= 5:
            return 2
        elif (bin - item) <= 7:
            return 1
        elif (bin - item) <= 9:
            return 0.9
        elif (bin - item) <= 12:
            return 0.95
        elif (bin - item) <= 15:
            return 0.97
        elif (bin - item) <= 18:
            return 0.98
        elif (bin - item) <= 20:
            return 0.98
        elif (bin - item) <= 21:
            return 0.98
        else:
            return 0.99
    return np.array([s(bin, item) for bin in bins])
\end{lstlisting}
\end{center}

\paragraph{Priority function and analysis}

The authors of FunSearch observed that several of the FunSearch heuristics
\textit{``assign[ed] items to least capacity bins only if the fit is very tight
after placing the item.''}~\cite{romera2024mathematical}. We verify this claim,
and provide a more precise characterization of their behavior.

\begin{figure}[tbh]
\resizebox{.5\linewidth}{!}
{
\begin{tikzpicture}[
    thick,
    >=stealth',
    dot/.style = {
      draw,
      fill = white,
      circle,
      inner sep = 0pt,
      minimum size = 4pt
    }
  ]
\begin{scope}[xscale=.2]
  \coordinate (O) at (0,0);
  \draw[->] (-0.4,0) -- (40,0) coordinate[label ={below:{\begin{tabular}{r}remaining \\space\end{tabular}}}] (xmax);
  \draw[->] (0,-0.1) -- (0,5) coordinate[label = {right:$priority$}] (ymax);

    \draw[blue,thick]      (0,4) -- (2,4) -- (2,3) -- (3,3)-- (3,2) -- (4,2)-- (5,2)-- (5,1) -- (7,1);
    \draw[dashed,thick]      (0,1) -- (40,1);
    \draw[thick]  (0.1,1) -- (-.3,1) node[left] {1};
    \draw[red,thick] plot[smooth] coordinates {(7.5,.1) (10,.3) (15,.43) (18,.53) (21.5,.58)};
    \draw      (22,.8) -- (40,.8);

    \draw (7,0.05) -- (7,-0.05) node[label = {below:$7$}] {};
    \draw (22,0.05) -- (22,-0.05) node[label = {below:$22$}] {};

    \draw[blue, dashed, <-] (2.3,3) -- (7,4) node[right] {"best fit"};
    \draw[red, dashed, <-] (15,.5) -- (18,4) node[right] {never};
    \draw[black, dashed, <-] (30,.9) -- (32,4) node[right] {first fit};

\end{scope}
\end{tikzpicture}}
\caption{Priority provided by \cTwelve. The scale of the $y$-axis is not linear
to highlight the difference of priority.\label{fig:visual_c12}}
\end{figure}
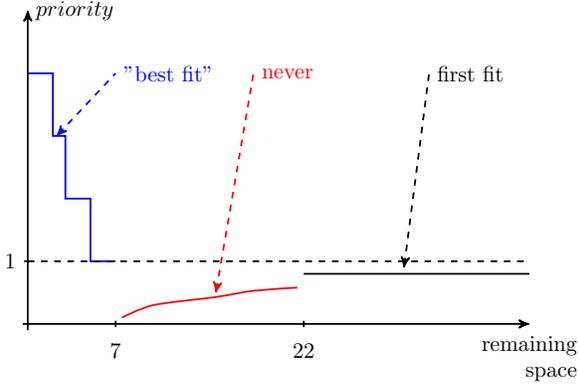

\cTwelve is the most easily interpretable heuristic in \romera.
It consists of a list of 11 if/then statements where conditions on the remaining bin capacity after placing the item determine the bin’s priority. 
One can notice that the priority of the bin is the highest when $bin - item$ is less
or equal to 7 while decreasing when $bin - item$ increases.
It means that if the remaining size of a bin after putting the item is below 7,
the bin will be selected while prioritizing the fullest bin among them. If no such
bin exists, the heuristic will select the first bin leaving more than 21
remaining size (line 23). 
Since empty bins are always available, bins that would leave between 7 and 21 remaining capacity are never
selected by the heuristic. 
Note that, in practice, Lines 11-20 of \cTwelve are never useful and can be removed. 
A graphical approximation of the algorithm’s behavior is provided in Figure~\ref{fig:visual_c12}.

\paragraph{Discussion on the performance}

\begin{figure}[tbh]
\centering
\includegraphics[width=0.5\linewidth]{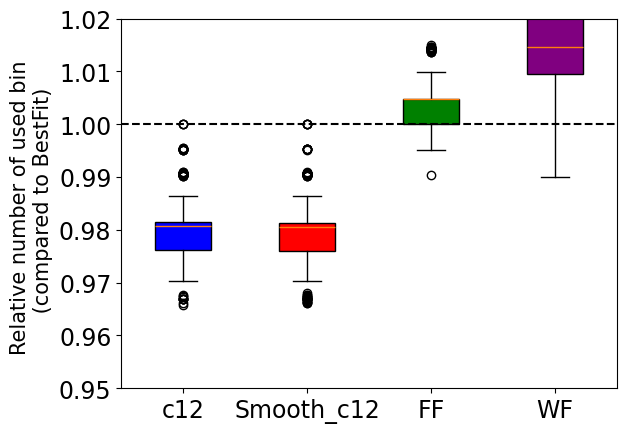}
\caption{Boxplots showing the heuristics performance relative to \bestfit over 1000 instances of the \unifDef distribution with 500 items and bin capacity 150.}
\label{fig:boxplot_uniform}
\end{figure}

\cTwelve can be simplified by {\em smoothing} its behavior: applying \bestfit to bins 
where the remaining space after placing the item would be less than 7 (if such bins exist), and otherwise applying
\firstfit to bins where the remaining space would be larger than 21. We refer to this heuristic as \smoothcTwelve. 
In border cases, these two algorithms may behave slightly differently:
 for instance, two bins leaving 4 and 5 remaining size will be treated
identically by \cTwelve, while \smoothcTwelve will prioritize the fullest one.
However, our experiments show that their performance, in terms
of the number of bins used, \revision{is identical on 68.8\% of the tested instances and differs by fewer than 2 bins on all instances (see Figure~\ref{fig:boxplot_uniform}).}

\begin{figure}[tbh]
\centering
\includegraphics[width=0.5\linewidth]{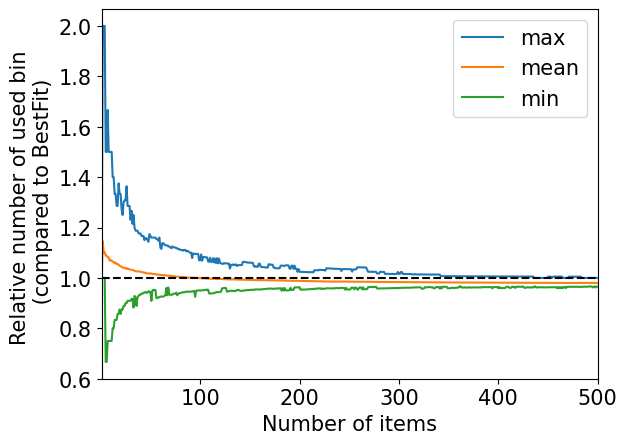}
\caption{Performance of \cTwelve relative to \bestfit over 1000 instances of the \unifDef distribution, as the number of items increases (bin capacity = 150).}
\label{fig:growing_c12}
\end{figure}

We hypothesize that the average 2.0\% gain of \cTwelve and \smoothcTwelve over \bestfit is due to
several combined factors:
\begin{itemize}
    \item (H1) \textbf{The large number of items to be scheduled.}
    We hypothesize that the performance gain arises when many items must be scheduled. 
    This is supported by Figure~\ref{fig:growing_c12}, which shows that \bestfit outperforms \cTwelve 
    when the number of items is small: for fewer than 90 items, the average performance of \cTwelve
     is lower than that of \bestfit.
    \item (H2) \textbf{The minimum item size imposed by the distribution.}
    Indeed, since all generated items are larger than 20, placing an item in a bin that leaves less than 20 units of 
    remaining capacity effectively {\em closes} that bin.
   Therefore, a strategy that avoids closing bins with {\em large} remaining capacity is advantageous when many items are expected, 
   as it increases the chance of closing the bins more efficiently  later.
We discuss (H2) in Section~\ref{sec:interesting} with other uniform
distributions by studying a parametrized version of \smoothcTwelve: \abff
(Algorithm~\ref{alg:abff}).
\end{itemize}

\subsection{Item sizes following a \weibDef distribution with bin capacity 100}
\label{sec:weibdef}

We now analyze the heuristics presented in \cite{romera2024mathematical} and
\cite{liu2024evolution} for the \weibDef distribution. The instances consist of
items with integer sizes drawn from a
 Weibull distribution with shape parameter $k = 3.0$ and scale parameter
$\lambda = 45$. Each bin has a capacity of 100.

\subsubsection{\cFourteen heuristic}

\paragraph{Heuristic generation}
The best performing heuristic for \weibDef by \romera is called \cFourteen
(Algorithm~\ref{alg:c14}). It was generated by evolving FunSearch on 5 bin
packing instances of 5000 items.

\begin{center}
 \begin{lstlisting}[language=Python,caption=\cFourteen~\cite{romera2024mathematical} for items of distribution \weibDef,label=alg:c14]
def priority_c14(item: float, bins: np.ndarray) -> np.ndarray:
    score = (bins - max(bins))**2 / item + bins**2 / item**2 + bins**2 / item**3
    score[bins > item] *= -1 
    score[1:] -= score[:-1]
    return score
\end{lstlisting}
\end{center}

\paragraph{Priority function and analysis}

\cFourteen is considerably less interpretable. Two aspects are particularly surprising: (i) the first
formula for the score function (line 2), as well as (ii) the final update of the
score function, which makes the score of a bin dependent on the score of one of its
neighbors (line 4).

\begin{figure}[th]
\begin{tikzpicture}
\begin{scope}
\begin{axis}[xmin=0,ymin=0,xlabel={Remaining bin capacity ($b$)},ylabel={Score (up until line 2)},ylabel near ticks,xlabel near ticks,clip=false,xticklabels={},yticklabels={},
]
    \addplot [
        ultra thick,
        magenta,
        domain=20:100,
        samples=100,
    ]
        {(x-100)^2 / 20 + x^2/20^2 + x^2 / 20^3};

    \addplot [
        ultra thick,
        blue,
        domain=50:100,
        samples=100,
    ]
        {(x-100)^2 / 50 + x^2/50^2 + x^2 / 50^3};
    \addplot[thick, samples=50, smooth,blue, dashed] coordinates {(50,0)(50,80)};

    \addplot[thick, samples=50, smooth,domain=0:6,magenta, dashed] coordinates {(20,0)(20,350)};
    \addplot[thick, samples=50, smooth,black, dashed] coordinates {(100,0)(100,80)};

    \node [magenta,thick] at (20,-10)   {$s_1$}; 
    \node [blue,thick] at (50,-10)   {$s_2$}; 
    \node [black,thick] at (100,-10)   {$c$}; 

    \node [magenta,thick] at (80,350)   {\tiny $f(s_1,b)=\frac{(b-c)^2}{s_1} + \frac{b^2}{s_1^2} + \frac{b^2}{s_1^3}$}; 
    \node [blue,thick] at (80,300)  {\tiny $f(s_2,b)=\frac{(b-c)^2}{s_2} + \frac{b^2}{s_2^2} + \frac{b^2}{s_2^3}$}; 
\end{axis} 
\end{scope}
\end{tikzpicture}
\caption{Plotting $f: s,b\mapsto \texttt{score}(s,b)$, where \texttt{score} is
the function from line 2 of \cFourteen. In these plots we use: $c=100$,
$s_1=20$, $s_2=50$. There is an inflexion point at
$\frac{c}{1+\frac{1}{s}+\frac{1}{s^2}}$.\label{fig:f}}
\end{figure}
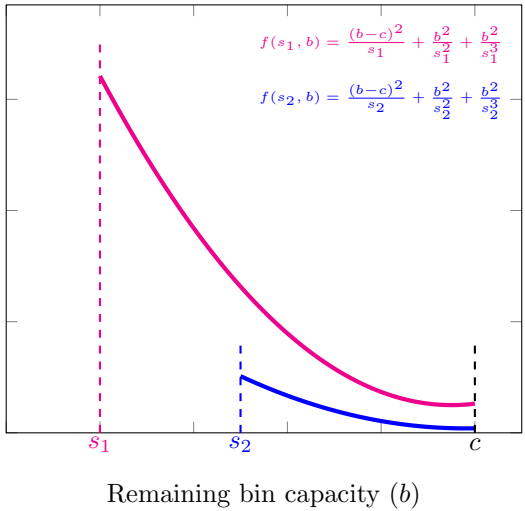

The score function defined on line 2 depends on the item size $s$, the maximum
bin capacity $c$, and the current remaining space $b$ in each bin to
determine each bin priority, with the formula:
\[f : (s,b) \mapsto \frac{(b - c)^2}{s} + \frac{b^2}{s^2} + \frac{b^2}{s^3}\]
Function $f$ is ilustrated in Figure~\ref{fig:f}. 
In theory, it is a decreasing function on $\left[s;\frac{c}{1+\frac{1}{s}+\frac{1}{s^2}}\right]$, which then slightly increases on $\left[\frac{c}{1+\frac{1}{s}+\frac{1}{s^2}};c\right]$. 
In practice, given that the distribution of item sizes $s$ contains few small values, and for large bin capacities $c$, the function $f$ is almost always decreasing.

Line 3 takes the additive inverse for each bin that is not a \textit{perfect-fit} (i.e. $b > s$), leaving only perfect-fit bins with a positive score.
Thus, up to line 3, the priority function can be interpreted as follows: if a \textit{perfect-fit} bin (i.e., a bin whose remaining capacity equals the item size) exists, it is selected;
otherwise, a new bin is opened.

Line 4 is arguably the most opaque part of the LLM-evolved heuristic: it
subtracts from each bin's score the score of the \emph{previous} bin in the implementation order.
We consider this to be the point at which the interpretability of the algorithm based solely on its code ends.
Beyond this point, our analysis is
experiment-based and required several sets of experiments to truly understand
the actual behavior of the heuristic. In the following, we present only
the successful ones.

\paragraph{Experimental analysis of \cFourteen's behavior}

\revision{Figure~\ref{fig:f} shows the shape of the score function $f$ for two item sizes ($s_1=20$ and $s_2=50$) to provide insight into the mechanism of \cFourteen.}  
After line 4 of the algorithm, a bin will have a high priority if its own $f$
value is low (i.e. it is almost empty) and/or its predecessor's $f$ value is
high (i.e. its predecessor is ``full-ish''). 
This observation prompted us to investigate the actual differences in behavior
between \cFourteen and \worstfit.

\begin{figure}[tbh]
\centering
\includegraphics[width=0.5\linewidth]{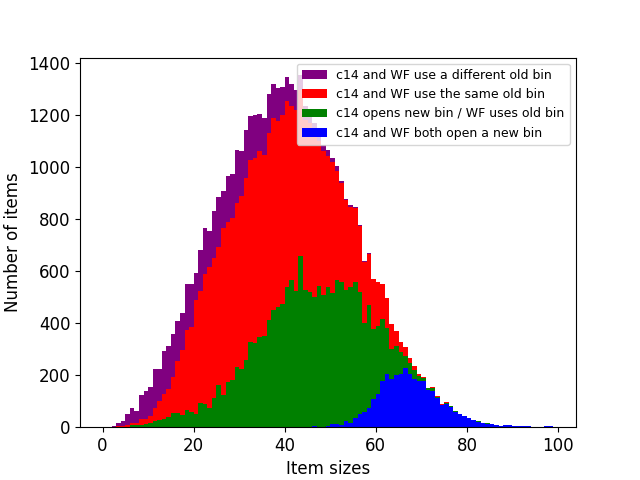}
\caption{Comparing the behavior item-per-items of \cFourteen and \worstfit on the \weibDef distribution with 50k items and bin capacity 100.}
\label{fig:hist}
\end{figure}

To this end, we conducted an item-by-item comparison between the decisions taken by \cFourteen and those that would have been selected by \worstfit.
In the following, we refer to an {\em old} bin as one that already contains 
at least  one item, and to a {\em new} bin as one that is empty.
Figure~\ref{fig:hist} shows, for each new item, the cases where both heuristics behave identically (both open a new bin, or both 
schedule the item in the same old bin) and the cases where they differ (selecting different old bins, or \cFourteen opening a 
new bin while \worstfit would not).
Note that if \cFourteen can place an item in an old bin, then \worstfit can as well, and will do so. 
Consequently, there is no case where \worstfit opens a new bin while \cFourteen uses an old one.
The main observations are as follow: when neither algorithm opens a new bin, they most often
 schedule the item in the same bin, resulting in similar behavior. 
 However, \cFourteen frequently opens a new bin when \worstfit does not, i.e., when a bin with sufficient remaining space exists.

\begin{figure}[tbh]
\centering
\subfloat[\cFourteen and \worstfit use \\a different old bin \label{fig:nuage1}]{\includegraphics[width=0.355\linewidth]{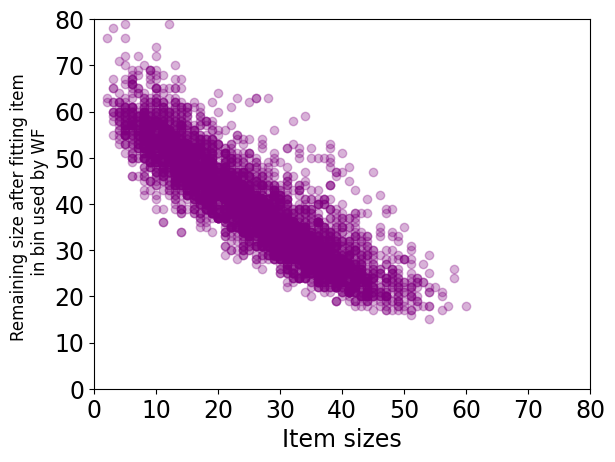}}
\subfloat[\cFourteen and \worstfit use \\the same old bin \label{fig:nuage2}]{\includegraphics[width=0.33\linewidth]{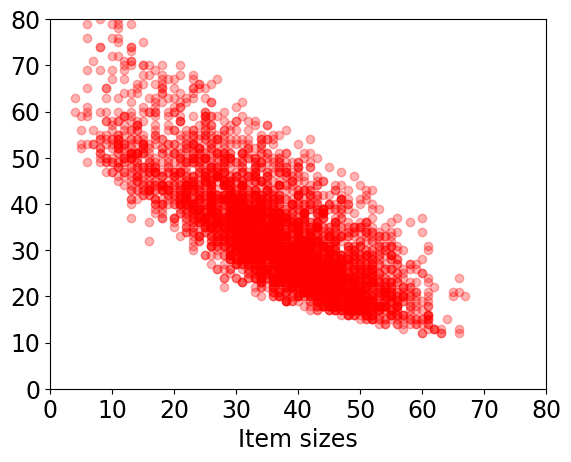}}
\subfloat[\cFourteen opens a new bin, \\ \worstfit uses an old bin \label{fig:nuage3}]{\includegraphics[width=0.33\linewidth]{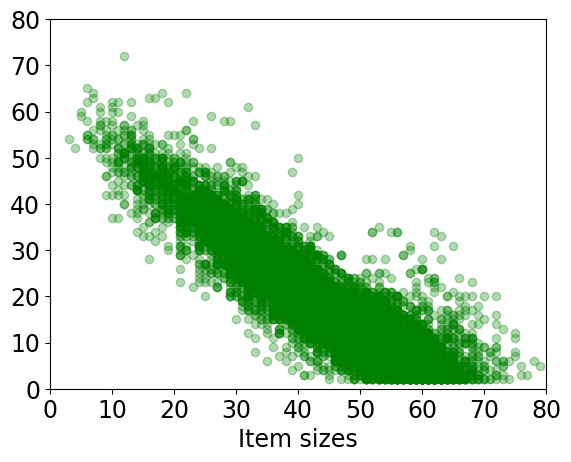}}
\caption{Remaining size in the bins used by \worstfit after fitting the item in a non-empty bin on the \weibDef distribution with 50k items and bin capacity 100.}
\label{fig:nuage}
\end{figure}

To better understand this divergence, we examine, in Figure~\ref{fig:nuage}, the remaining space in the bins chosen by \worstfit whenever \cFourteen instead opens a new bin.
Figure~\ref{fig:nuage} illustrates \cFourteen’s behavior from a different perspective:
in practice, when \worstfit would schedule an item into a bin whose remaining space is less 
than the item size plus a certain threshold (approximately 20), \cFourteen instead opens a new bin.

Finally, based on all these evaluations, we propose the following practical interpretation of
\cFourteen's general behavior: apply \bestfit if a perfect-fit bin exists; otherwise, if there exists a bin whose remaining space exceeds the item size by more than 20,
 apply \worstfit; if neither condition holds, open a new bin.
Note that, unlike for \cTwelve, this interpretation differs non negligibly from the
actual behavior of \cFourteen (see Figure~\ref{fig:nuage}).

\paragraph{Discussion on the performance}

\begin{figure}[tbh]
\centering
\includegraphics[width=0.5\linewidth]{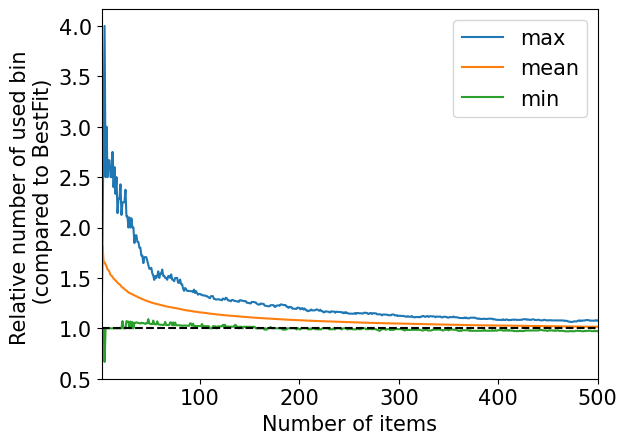}
\caption{Average performance of \cFourteen relative to \bestfit over 1000 instances of the \weibDef distribution, as the number of items increases (bin capacity = 100).}
\label{fig:growing_c14}
\end{figure}

In practice \cFourteen appears to be a more opaque implementation of the same
underlying principle as \cTwelve, which relies on two key hypotheses: (H1) the presence of a large number of items to be scheduled, 
and (H2) a distribution containing few elements smaller than a given threshold.
As in the \unifDef case, (H1) can be verified by examining the relative performance
of \cFourteen compared to \bestfit as the number of items increases
(Figure~\ref{fig:growing_c14}). In this case, the performance gap for small
numbers of items is even more pronounced.

\subsubsection{EoH heuristic}
\cite{liu2024evolution} extended the work of \romera and developed the EoH
heuristic for the \weibDef distribution. Their methodology is largely similar,
differing mainly in the evolutionary framework employed. \revision{The authors
used the same five instances than \romera to evolve their heuristic.} Their
resulting algorithm is presented in Algorithm~\ref{alg:eoh}.

 \begin{lstlisting}[language=Python,caption=The priority function evolved with EoH~\cite{liu2024evolution} for items from distribution \weibDef,label=alg:eoh]
def priority_EoH(item: float, bins: np.ndarray) -> np.ndarray:
    diff = bins-item # remaining capacity
    exp = np.exp(diff) # exponent term
    sqrt = np.sqrt(diff) # square root term
    ulti = 1-diff/bins # utilization term
    comb = ulti * sqrt # combination of utilization and square root
    adjust = np.where(diff > (item * 3), comb + 0.8, comb + 0.3) # hybrid adjustment term to penalize large bins
    hybrid_exp = bins / ((exp + 0.7) *exp) # hybrid score based on exponent term
    scores = hybrid_exp + adjust # sum of hybrid score and adjustment
    return scores
\end{lstlisting}

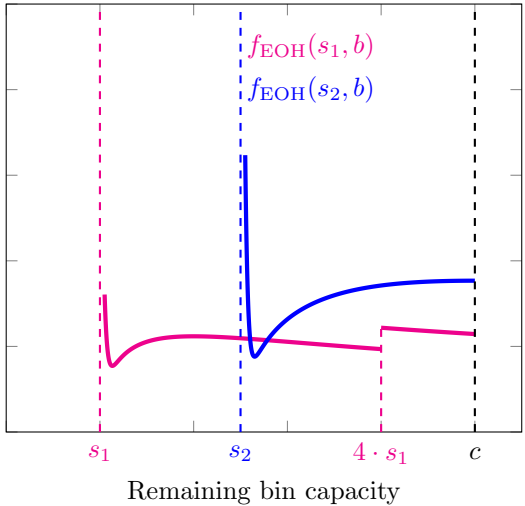
\begin{figure}[th]
\begin{tikzpicture}
\begin{axis}[xmin=0,ymin=0,ymax=10,xlabel={Remaining bin capacity},ylabel={Score},ylabel near ticks,xlabel near ticks,clip=false,xticklabels={},yticklabels={},
]
    \addplot [
        ultra thick,
        magenta,
        domain=21:80,
        samples=1000,
    ]
        {x / ((e^(x-20) + 0.7)*e^(x-20)) + (1- (x-20)/x) * sqrt(x-20)};
        
     \addplot [
        ultra thick,
        magenta,
        domain=80:100,
        samples=1000,
    ]
        {0.5 + x / ((e^(x-20) + 0.7)*e^(x-20)) + (1- (x-20)/x) * sqrt(x-20)};

    \addplot [
        ultra thick,
        blue,
        domain=51:100,
        samples=1000,
    ]
        {x / ((e^(x-50) + 0.7)*e^(x-50)) + (1- (x-50)/x) * sqrt(x-50)};
    \addplot[thick, samples=50, smooth,blue, dashed] coordinates {(50,0)(50,10)};

    \addplot[thick, samples=50, smooth,domain=0:6,magenta, dashed] coordinates {(20,0)(20,10)};
    \addplot[thick, samples=50, smooth,domain=0:6,magenta, dashed] coordinates {(80,0)(80,2.5)};
    \addplot[thick, samples=50, smooth,black, dashed] coordinates {(100,0)(100,10)};

    \node [magenta,thick] at (20,-5)   {$s_1$}; 
    \node [magenta,thick] at (80,-5)   {$4\cdot s_1$}; 
    \node [blue,thick] at (50,-5)   {$s_2$}; 
    \node [black,thick] at (100,-5)   {$c$}; 

    \node [magenta,thick] at (65,90)   {$f_{\text{EOH}}(s_1,b)$}; 
    \node [blue,thick] at (65,80)  {$f_{\text{EOH}}(s_2,b)$}; 
\end{axis} 
\end{tikzpicture}
\caption{Plotting $s,b\mapsto f_{\text{EOH}}(s,b)$ when the bin remaining capacity $b$ varies. In these plots, we use: $c=100$, $s_1=20$, $s_2=50$.}
\label{fig:f2}
\end{figure}

In practice this score function $f_{\text{EOH}}$ is less opaque than that of
\cFourteen as it depends solely on the item size $s$ and the remaining bin capacity $b$, rather than on adjacent bins:
\begin{align*}
f_{\text{EOH}} : (s,b) \mapsto &\frac{b}{(e^{b-s}+0.7) e^{b-s}} + \left (1- \frac{b-s}{b} \right ) \sqrt{b-s} + (0.5\text{ if }b > 4s)
\end{align*}
However it remains largely hard to explain, particularly given the case depending on whether $b>4s$ (line 7), which only occurs when the item size is small enough.
Figure~\ref{fig:f2} provides a visual representation of this function. 

\subsection{Global discussion on interpretability}

\romera claim that one of the main benefits of the LLM-evolved strategies is
their interpretability: {\em ``Beyond being an effective and scalable strategy,
discovered programs tend to be more interpretable than raw solutions."}
Interpretability goes beyond the mere ability to {\em read} an algorithm's behavior. 
It is understood as enabling humans to grasp the motivations underlying this behavior 
and to generate new knowledge from it.
This distinction between readability and interpretability echoes a broader issue in AI research: 
human-readable outputs do not necessarily translate into human-understandable reasoning or transferable knowledge.  
Specifically, {\em ``a method is
interpretable if a user can correctly and efficiently predict the method’s
results''}~\cite{kim2016examples}.

The question of interpretability can then be rewritten as follows: by looking at
the algorithms produced (\cTwelve,\cFourteen or \eoh), would someone have been
able to predict an improved performance compared to \bestfit? We have shown that
the answer for \cFourteen and \eoh is no.

While \revision{the performance of} \cTwelve was relatively easy to interpret, understanding the performance
of \cFourteen required scheduling expertise, multiple attempts (we only show the
successful path) and several rounds of experimentation. 
Ultimately, our interpretation of \cFourteen
did not come from the description of the score function, but from
experimental observations of the algorithm's behavior. For instance, the threshold value of 21
observed in Figure~\ref{fig:nuage} is not apparent anywhere in the code. {\bf In
that sense, our experimental analysis was no different from what would have been
required to interpret a fully black-box algorithm, \revision{contradicting the fact that
the specificity of the LLM was what made it easier to interpret.}}
The fact that subsequent work has not attempted to reuse the {\em knowledge}
supposedly created by these LLM-evolved heuristics does not bode well for their
interpretability.

For these reasons, we find that \cFourteen is not interpretable in the sense
defined by \romera and Kim et al.~[\citeyear{kim2016examples}].



We conclude this section by noting that the approach of
instantiating a parameterized priority function is not new in scheduling as a
way to {\em interpret} algorithmic performance. For instance, a similar approach was used to solve the
problem of scheduling parallel tasks with release times, aiming to minimize the mean
bounded slowdown~\cite{carastan2017obtaining}. The result states that tasks should be sorted
according to the following priority function $f_1: (r,s,n)\mapsto \log_{10}(r) \cdot n + 8.7 \cdot
10^2 \cdot \log_{10}(s)$, where $r$ denotes the release time of the task, $n$ the
number of nodes, and $s$ its size. As with \cFourteen, this expression is readable and allows for some intuition,
but it does not mean that it is (easily) interpretable.

\section{Assessing the scientific value of LLM-evolved ideas}
\label{sec:interesting}

Having discussed interpretability, we now turn to the related question: are the ideas 
generated by the LLM-based genetic frameworks scientifically interesting? 

Indeed, the second main claim of \romera is that the use of LLMs in FunSearch \textit{``represents the first
discoveries made for established open problems using LLMs''}. Specifically for
bin packing, FunSearch is reported to have {\em `` discover[ed] a program that corresponds to a more
efficient heuristic for online bin packing.''}. This claim is supported by
performance comparisons against baseline methods, and they further emphasize that these
\textit{``discoveries are truly new''}.

As noted in the previous sections, these LLM-based genetic studies do not address the canonical open problem of {\em online bin packing}.
Instead they focus on a narrower and previously unstudied instance of the problem: {\em online
bin packing with items sizes drawn from specific parameterizations of the uniform or Weibull
distributions}. 
In addition, subsequent work~\cite{sim2025beyondhype} has shown that
the generated solutions fail to generalize to other parameter settings of the uniform or
Weibull distributions.
In our view, one way to assess the scientific value of these heuristics is to examine whether 
they can serve as a basis for deriving new ones applicable to other instances of the uniform or Weibull distributions.

\subsection{Can we generalize the idea behind \cTwelve, \cFourteen, and \eoh?}
A naive inspection of \cTwelve suggests a solution with too many parameters to be
directly instantiated. However, we have shown that its behavior can be approximated
by a two-parameter heuristic: apply \bestfit if there exists a bin where the
remaining capacity would be less than 7, otherwise apply \firstfit to bins with remaining capacity greater than 21.

The main idea underlying \cFourteen and \eoh is the use of an analytical priority function.
In our view, however, the parameterizations of these
functions are too complex to serve as a foundation for constructing new ideas. 
This observation is consistent with the fact that subsequent studies have continued to rely on LLM-evolved heuristics
rather than building on the underlying {\em new idea} itself.

\subsection{Generalization of our analysis}
Our analysis of the LLM-evolved heuristics suggests that, at a fundamental level, their
performance relies on the same behavioral principle, which can be generalized as follows:
\begin{enumerate}
    \item An item of size $s$ is allocated to the best-fitting bin if the fit is sufficiently tight, 
    i.e., to bins whose remaining capacity is less than $s + a$ for a threshold $a$.
    \item Otherwise, it is allocated to a bin whose remaining capacity is greater than
$s+b$ for a threshold $b$, using any of the {\em
baseline} heuristics. 
    \item If no such bin exists, the item is placed into a new bin.
\end{enumerate}
In the remainder, we denote by \abff (resp. \abwf, \abbf) this heuristics using \firstfit
(resp. \worstfit, \bestfit) as the baseline strategy for the second step.
Their priority function, described in Algorithms~\ref{alg:abff},~\ref{alg:abbf}, and ~\ref{alg:abwf}, differ only at line~8 of their implementation.

\begin{lstlisting}[language=Python,label=alg:abff,caption=Priority function for \abff for thresholds $a$ and $b$,escapeinside={(*@}{@*)},linebackgroundcolor={\ifnum\value{lstnumber}>6\ifnum\value{lstnumber}<9\color{black!10}\fi\fi}]
def priority_abFF(item: float, bins: np.ndarray) -> np.ndarray:
    def s(bin, item):
        if bin <= item + (*@  \textcolor{blue}{a}  @*):
            return capacity - bin + 1
        elif bin < item + (*@  \textcolor{blue}{b}  @*):
            return -2
        else:
            return 1 
    return np.array([s(bin, item) for bin in bins])
\end{lstlisting}
\begin{lstlisting}[language=Python,label=alg:abbf,caption=Priority function for \abbf for thresholds $a$ and $b$,escapeinside={(*@}{@*)},linebackgroundcolor={\ifnum\value{lstnumber}>6\ifnum\value{lstnumber}<9\color{black!10}\fi\fi}]
def priority_abBF(item: float, bins: np.ndarray) -> np.ndarray:
    def s(bin, item):
        if bin <= item + (*@  \textcolor{blue}{a}  @*):
            return capacity - bin + 1
        elif bin < item + (*@  \textcolor{blue}{b}  @*):
            return -2
        else:
            return 1/(bin - item)
    return np.array([s(bin, item) for bin in bins])
 \end{lstlisting}
 \begin{lstlisting}[language=Python,label=alg:abwf,caption=Priority function for \abwf for thresholds $a$ and $b$,escapeinside={(*@}{@*)},linebackgroundcolor={\ifnum\value{lstnumber}>6\ifnum\value{lstnumber}<11\color{black!10}\fi\fi}]
def priority_abWF(item: float, bins: np.ndarray) -> np.ndar!ray:
    def s(bin, item):
        if bin <= item + (*@  \textcolor{blue}{a}  @*):
            return capacity - bin + 1
        elif bin <= item + (*@  \textcolor{blue}{b}  @*):
            return -2
        elif bin == capacity:
            return -1
        else:
            return -1/(bin-item)    
    return np.array([s(bin, item) for bin in bins])
\end{lstlisting}

As discussed earlier, the performance of this class of heuristics can be explained by two main factors:
(H1) a low number of elements whose sizes are below a given threshold; and (H2) a large total number of items to be scheduled.

\subsection{Performance of \abff, \abwf, \abbf}

\begin{figure}[tbh]
\centering
\subfloat[\abff on \unifDef.\\ Bin capacity = 150, number of items = 500.\\
Minimum for $a = 5$ and $b = 24$, relative ratio: $0.979$.\label{fig:abff}]{
\includegraphics[width=0.5\linewidth]{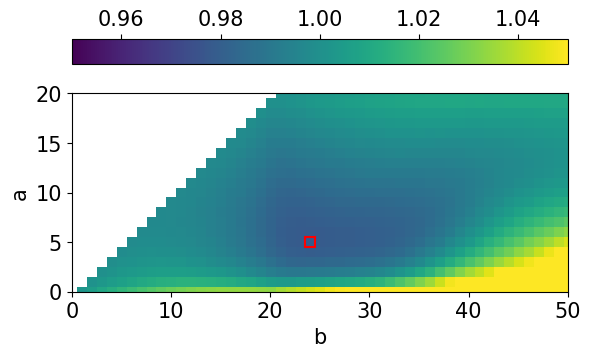}}
~~\subfloat[\abwf on \weibDef.\\ Bin capacity = 100, number of items = 5000.\\
Minimum for $a = 1$ and $b = 21$, relative ratio: $0.967$.\label{fig:abwf}]{
\includegraphics[width=0.5\linewidth]{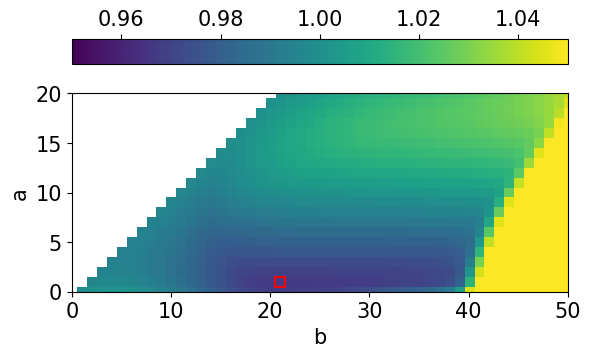}}
\caption{Heatmap showing the relative number of bins used by \abff/\abwf for various values of $a$ and $b$, compared to \bestfit, over 100 instances of different distributions.}
\label{fig:9}
\end{figure}

\begin{figure}[tbh]
\centering
\subfloat[\ema{\textsc{Uniform}(20,100)}, BC = 150 \label{fig:boxplot_uniform_20_100_150}]{\includegraphics[width=0.33\linewidth]{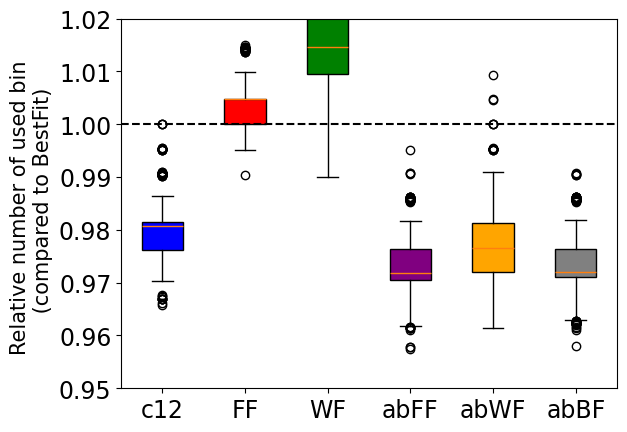}}
\subfloat[\ema{\textsc{Uniform}(40,120)}, BC = 250 \label{fig:boxplot_uniform_40_120_250}]{\includegraphics[width=0.33\linewidth]{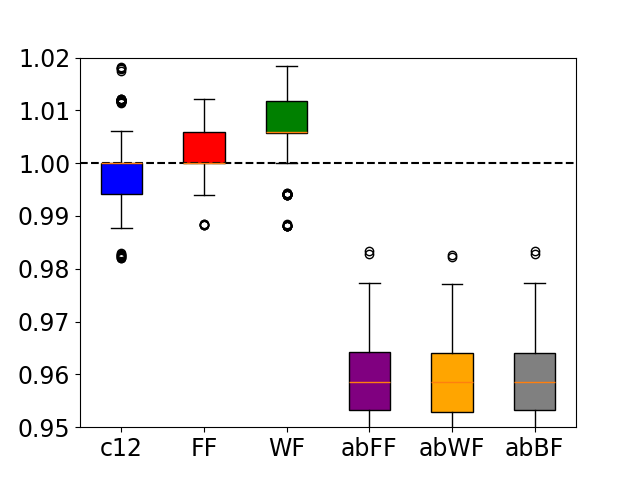}}
\subfloat[\ema{\textsc{Uniform}(60,140)}, BC = 350 \label{fig:boxplot_uniform_60_140_350}]{\includegraphics[width=0.33\linewidth]{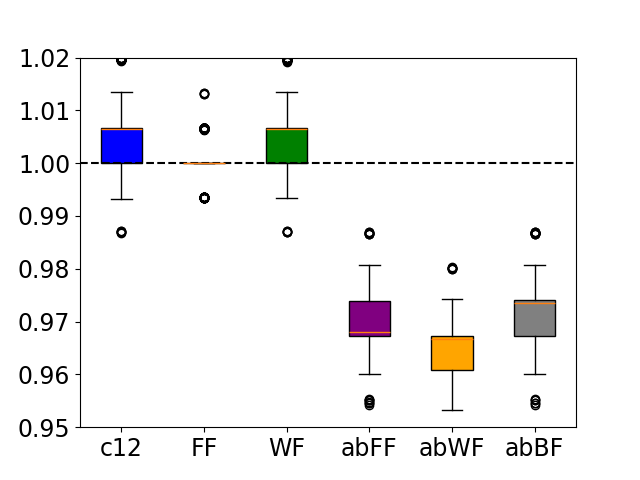}}

\subfloat[\ema{\textsc{Weibull}(3.0,45)}, BC = 100 \label{fig:boxplot_weibull_3_45_100}]{\includegraphics[width=0.33\linewidth]{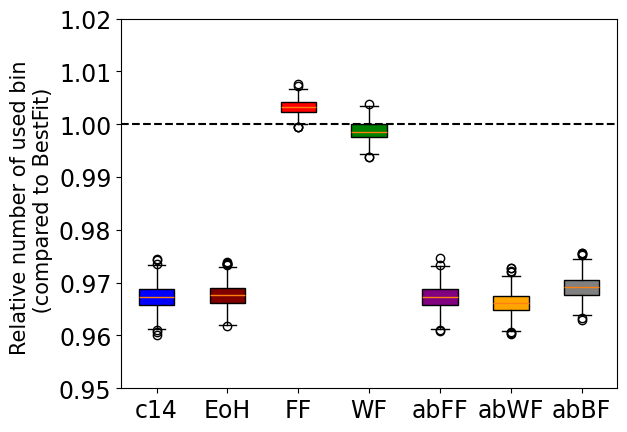}}
\subfloat[\ema{\textsc{Weibull}(5.0,60)}, BC = 300 \label{fig:boxplot_weibull_5_60_300}]{\includegraphics[width=0.33\linewidth]{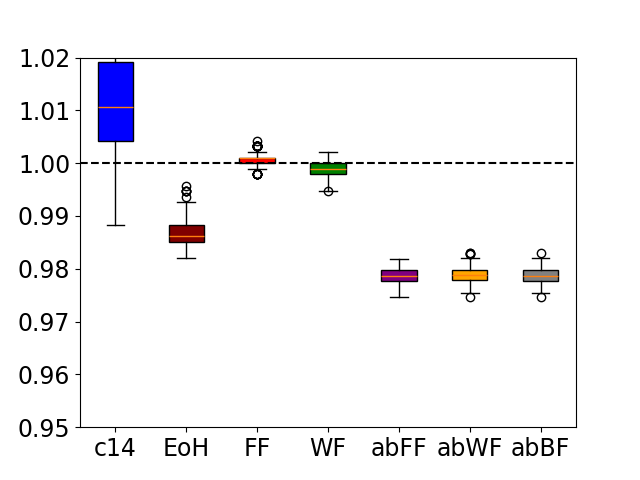}}
\subfloat[\ema{\textsc{Weibull}(7.0,75)}, BC = 500 \label{fig:boxplot_weibull_7_75_500}]{\includegraphics[width=0.33\linewidth]{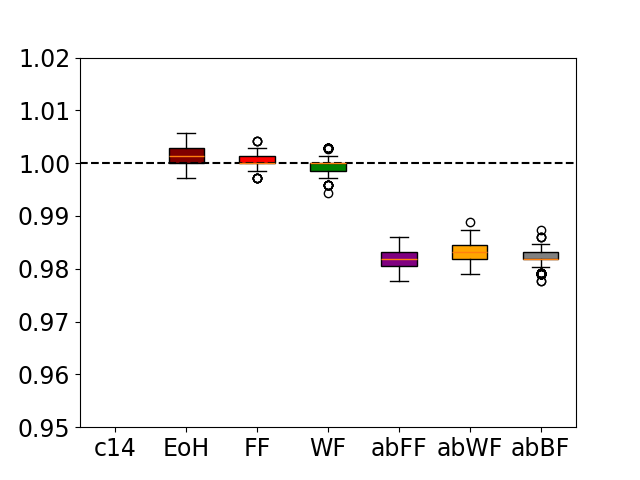}}
\caption{Performance of \abff, \abbf, and \abwf across various distributions and
various Bin Capacities (BC). \revision{Each boxplot represents 1000 instances.} For \textsc{Weibull}(7.0,75), the performance of
\cFourteen is too degraded to be visible in the graph.}
\label{fig:bigperf}
\end{figure}

In this section we evaluate the performance of our parametrizable heuristics across
various distributions, including those studied in \romera.
\revision{Figure~\ref{fig:9} shows the performance of \abff and \abwf for several combinations of
parameters $(a,b)$ averaged over 100 runs.
Figure~\ref{fig:bigperf} shows the performance of the heuristics when selecting the best combinaisons
of parameters $(a,b)$ over 1000 runs.}

\paragraph{\unifDef.}

Figure~\ref{fig:abff} shows the relative number of bins used by \abff compared to
\bestfit for various values of $a$ and $b$, using the same \unifDef distribution on which \cTwelve was trained.
Several combinations of parameters $(a,b$) achieve comparable performance,
highlighting the robustness of the heuristic with respect to parameter selection.
For \unifDef, by selecting the best combination, $a=5$ and $b=24$, \abff (the
heuristic most similar to \cTwelve) \revision{has comparable performance to \cTwelve. It outperforms \cTwelve (resp. \bestfit) by 0.71\%
(resp. 2.71\%) on average and 0.88\% (resp. 2.82\%) on median. Complete
statistical data is available in Fig.~\ref{fig:boxplot_uniform_20_100_150}.}

\paragraph{\weibDef.}

Figure~\ref{fig:abwf} shows the relative number of bins used by \abwf compared
to \bestfit for various values of $a$ and $b$, using the same \weibDef
distribution on which \cFourteen was trained.
Again, several parameter combinations achieve similar performance. The best
performance is obtained for \abwf with parameters $a=1$ and $b=21$, with \revision{performance similar to those of \cFourteen. \abwf
outperforms \cFourteen (resp. \eoh, \bestfit) by 0.12\%
(resp. 0.15\%, 3.39\%) on average and 0.11\% (resp. 0.15\%, 3.38\%) on median. Complete
statistical data is available in Fig.~\ref{fig:boxplot_weibull_3_45_100}.}

\paragraph{Other uniform and Weibull distributions}

Figure~\ref{fig:bigperf} evaluates the performance of \abff, \abbf and \abwf
compared to the baseline strategies and the LLMs-evolved heuristics on the
\unifDef and \weibDef distributions, as well as on additional distributions.
\revision{Note that this is not meant to be a systematic review of performance of the
the ab-heuristics, hence the scope is quite limited. We used the shape (Uniform
and Weibull) used by the LLM literature and simply changed the parameters via an
incremental exploration. We selected values to have a variety of distributions
in sizes, as well as in the number of items that we can fit in a bin. We did not
experiment with other distributions than the ones ploted here, although the
interested readers can use our code to experiment more.}

The best-performing parameters $a$ and $b$ for \abff, \abbf, and \abwf, were
selected separately for each problem instance.

Let \abprio denote the best performing among the three heuristics \abff, \abbf,
and \abwf. \abprio outperforms all baselines and LLMs-evolved heuristics
considered, even on the data distributions on which they were trained.

Our experiments confirm that \cTwelve, \cFourteen, and other LLM-evolved
heuristics are not robust to variations in the data distribution, as previously demonstrated in detail \cite{sim2025beyondhype}. This result is not surprising: these
heuristics are highly optimized for a single distribution—just as \abprio is
with a fixed choice of parameters.

\begin{figure}[tbh]
\subfloat[\abff ($a=5$, $b=24$),\\ \unifDef, bin capacity = 150 \label{fig:growing_abff}]{\includegraphics[width=0.5\linewidth]{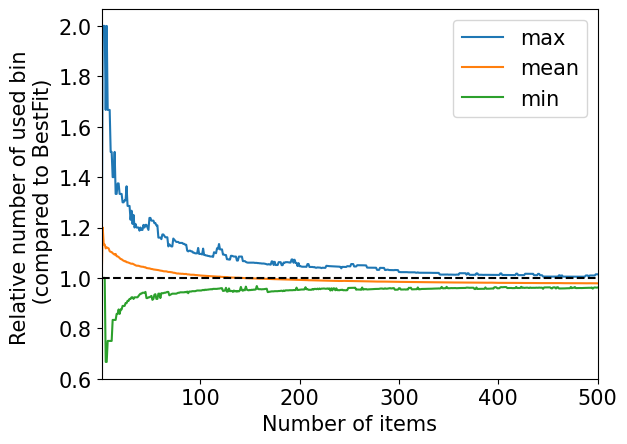}}
\subfloat[\abwf ($a=1$, $b=21$),\\ \weibDef, bin capacity = 100 \label{fig:growing_abwf}]{\includegraphics[width=0.5\linewidth]{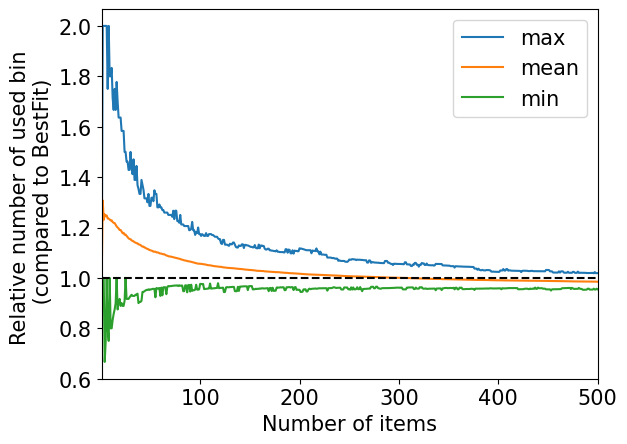}}
\caption{Average performance of some \abprio heuristics compared to \bestfit over 1000 instances of the selected distribution as the number of item increases. 
\label{fig:growing_abprio}}
\end{figure}

\abprio, however, generalizes the core idea easily and effectively to
distributions that exhibit a lower bound on item sizes and contain a large number of items.
It nevertheless suffers from the same limitation as \cTwelve and \cFourteen when the number of
items is small: the performance can degrade to as much as $c / (c-b)$ times the
optimal\footnote{This can be shown by considering an instance where all arriving
items are smaller than $b$, such that they would fill the bins up to $c-b$, then
the next item would open a new bin. Note that if $b<x$ the distribution minimum,
in this case the ratio would be $c/(c-x)>c/(c-b)$ and the result holds.}, where
$c$ is the bin capacity and $b$ the second parameter.
Notably, for \weibDef, \abwf converges much faster to comparable performance level than
\cFourteen (Figure~\ref{fig:growing_abwf}).

Finally, it is worth noting that all our experimental runs,
including the meta-tuning of \abprio, can be completed within a few minutes of
execution time on a personal laptop, contrasting sharply with the thousands or even millions of
LLMs queries required to generate the LLMs-evolved heuristics. Moreover, if
the community were to pursue this line of work, we expect that
performance could be further improved, from a computational complexity perspective, by employing 
dichotomic search for parameter tuning, as an example.

\subsection{Concluding Discussions}

In their work, \romera proposed using LLM-based solutions in the manner we have adopted: 
{\em FunSearch suggests a solution, which is examined by
researchers, who may note features of interest.}
In this sense, our study seems to support the process they envisioned, in which
LLMs assist researchers in exploring and refining new scientific insights.

Nevertheless, a broader question remains: can LLMs truly
contribute to solving {\em established open problems}, as claimed by Romera et al.,
thereby justifying their role in scientific discovery?
Indeed, their paper belongs to a broader research effort
aiming to demonstrate meaningful applications of LLMs—specifically, 
cases where the resulting advances would not have been achievable without machine involvement.
Such claims can only be supported through a rigorous validation protocol 
applied to a problem recognized as both challenging and significant within the scientific community.

\paragraph{On the absence of related work}

A key limitation of \romeras claims lies in the absence of prior studies on
the online bin packing problem where item sizes follow a \unifDef or
\weibDef distribution. This lack of related work raises several possible explanations.

One possibility is that no one had previously asked this particular question because it was not considered particularly important.
If so, the surge of interest in these bin packing instances since 2023—mostly driven by the 
narrative of LLM-based discovery—may reflect an artificial increase in their perceived importance 
rather than genuine theoretical significance.

Another explanation is that the problem, although perhaps already examined, 
was solved but never formally published.
If that were the case, it would raise an interesting contrast: 
human-generated solutions of this kind were possibly deemed too simple or 
insufficiently significant to merit publication, while similar LLM-generated results 
have recently led to a proliferation of preprints and publications~\cite{liu2024evolution,dat2025hsevo,ye2024reevo,chen2024qube}.
This contrast invites reflection on the expectations the research community places on AI-generated findings.

Finally, one cannot entirely rule out the possibility that relevant work does exist but has escaped our review.
However, if the considered problems were perceived as significant, it would likely have achieved at least minimal visibility in the literature.

Taken together, these observations suggest that the absence of related studies is 
more plausibly explained by a lack of perceived importance than by genuine novelty or difficulty.

\paragraph{LLMs for ``mathematical discovery”}

As discussed earlier, in the case of the online bin packing problem, the LLM-evolved heuristics
show limited interpretability even for domain experts. Our understanding of their behavior largely relied on experimental observation,
much like one would study a black-box heuristic.
By comparison, the \abprio meta-heuristic is more interpretable, more efficient, and more generalizable.
Its underlying rationale is relatively straightforward once the properties 
of the distribution and the large number of elements are considered.

It remains conceivable that knowing in advance that a better solution than \bestfit
 existed helped us identify it.
A central challenge of scientific research lies precisely in the 
uncertainty of whether better solutions exist at all. 
Having access to such knowledge, or to an oracle that provides it, 
would undoubtedly be a major advantage.
Yet, this advantage does not reflect genuine insight but rather a form of informed exploration.

Indeed, the patterns we identified are not explicitly encoded in the generated heuristics but emerged indirectly
from behavioral analysis. 
This indicates that the contribution of LLMs lies in empirical optimization rather than in genuine conceptual innovation.
Consequently, the case of online bin packing provides insufficient evidence to claim 
that LLMs have produced insights amounting to new \emph{mathematical discovery}.

\paragraph{On \romeras title}

The expression ``mathematical discovery” in \romeras title is conceptually broad and scientifically ambiguous. 
It is not clear at what point an incremental improvement qualifies as a ``mathematical discovery.” 
This ambiguity is particularly relevant in domains where empirical performance improvements and theoretical insights are tightly coupled but not equivalent.

From a technical perspective, \romeras contribution is valuable.
Extending the search operators of genetic algorithms with LLMs is an original idea that is worth investigating. 
Automatically generating heuristics that perform competitively with standard baselines on trained datasets—without domain-specific knowledge—can prove useful in specific applied settings. 
However, qualifying such generated heuristics as a ``mathematical discovery” requires a strong validation protocol and domain expertise.
Based on our analysis, these conditions do not seem to hold for the online bin packing problem.

It is noteworthy that subsequent studies~\cite{liu2024evolution,dat2025hsevo,ye2024reevo} that refined genetic frameworks using LLMs adopted the same validation protocol for their generated heuristics—comparing performance against \bestfit on the \weibDef distribution. 
Yet, none of these works claimed to have made a “mathematical discovery” when surpassing \cFourteen. 

This raises questions about whether \romeras title was intended more as an attention-oriented phrasing than as a precise scientific statement.
A title that overstates its claims can influence citation practices and public perception in ways that may not reflect the actual nature of the work. 
It would be informative to understand how this title went through the peer-review process.
Open or transparent review processes could help mitigate similar issues by promoting accountability and constructive discussion about scientific publications.


\section{Conclusion}
\label{sec:conclusion}

This study investigated the interpretability and scientific value of heuristics 
evolved through LLM-based genetic frameworks, focusing on the online bin packing problem.
Despite the claim of \emph{mathematical discovery}, our analysis of their behavior shows that these heuristics largely rediscover known patterns under narrow experimental conditions.
These patterns can be formalized into an interpretable and more efficient two-parameter algorithm that can be easily tuned across different distributions.
This abstraction not only generalizes beyond the LLM-evolved heuristics but also shows that their apparent novelty stems primarily from 
reparameterizing familiar strategies rather than introducing new algorithmic concepts.
These results suggest that the LLM-based frameworks did not develop a conceptual understanding of the problem but rather discovered effective strategies through stochastic search and empirical feedback,
similar to a generic hyper-heuristic.

Overall, LLMs appear as another approach for extending the capabilities of genetic algorithms and broadening the search space of hyper-heuristics.
In the case of online bin packing, these benefits remain primarily empirical and do not yet demonstrate a capacity for autonomous theoretical insight or abstraction.
A convincing validation of such a capability would require applying LLM-based
 methods to significant, well-established problems and showing that the resulting solutions achieve demonstrable
 improvements over the best results in the existing literature, rather than focusing on previously
unexplored instances.

\section*{Acknowledgments}

This work was supported by funding managed by the French National Research Agency (ANR) under the France 2030 program through the DAIMOS project (grant ANR-25-EXNU-0002). 

\section*{Usage of LLM}
The Writefull proofreading software incorporated into Overleaf has been use to improve some of the writing of this paper.

\bibliographystyle{ACM-Reference-Format}
\bibliography{biblio}

\end{document}